\pdfoutput=1

\documentclass[11pt]{article}

\usepackage{acl}

\usepackage{times}
\usepackage{latexsym}

\usepackage[T1]{fontenc}

\usepackage[utf8]{inputenc}

\usepackage{microtype}
\usepackage{caption}
\usepackage{subcaption}
\usepackage{graphicx}
\usepackage{hyperref}
\usepackage{float}

\title{Developmental Negation Processing in Transformer Language Models}

\author{Antonio Laverghetta Jr. \and John Licato \\
  Advancing Machine and Human Reasoning (AMHR) Lab \\
  Department of Computer Science and Engineering \\
  University of South Florida \\
  Tampa, FL, USA \\
  \texttt{\{alaverghett,licato\}@usf.edu} \\}

\begin{document}
\maketitle
\begin{abstract}
Reasoning using negation is known to be difficult for transformer-based language models. While previous studies have used the tools of psycholinguistics to probe a transformer's ability to reason over negation, none have focused on the types of negation studied in developmental psychology. We explore how well transformers can process such categories of negation, by framing the problem as a natural language inference (NLI) task. We curate a set of diagnostic questions for our target categories from popular NLI datasets and evaluate how well a suite of models reason over them. We find that models perform consistently better only on certain categories, suggesting clear distinctions in how they are processed.\footnote{Code and data to reproduce our experiments can be found on Github: \href{https://github.com/Advancing-Machine-Human-Reasoning-Lab/negation-processing-ACL-2022}{https://github.com/Advancing-Machine-Human-Reasoning-Lab/negation-processing-ACL-2022}}

\end{abstract}

\section{Introduction}

Negation is an important construct in language for reasoning over the truth of propositions \cite{10.1093/aristotelian/44.1.127}, garnering interest from philosophy \cite{Horn1989-HORANH}, psycholinguistics \cite{zwaan2012experiential}, and natural language processing (NLP) \cite{morante2021recent}. While transformer language models (TLMs) \cite{NIPS2017_3f5ee243} have achieved impressive performance across many NLP tasks, a great deal of recent work has found that they do not process negation well, and often make predictions that would be trivially false in the eyes of a human \cite{rogers-etal-2020-primer,ettinger-2020-bert,laverghetta-jr-etal-2021-transformer}.

In developmental psychology, there has likewise been a great deal of interest in how a child's ability to comprehend negation emerges in the early years of life \cite{nordmeyer2013measuring,nordmeyer2018early,reuter2018getting,grigoroglou2019toddlers}. Unlike in NLP, which typically treats negation as representing a single monolithic competency, this research has long understood that there are many kinds of negation used in everyday interactions \cite{bloom1970language,pea1982origins}. This ranges from using negation to express a child's rejection of something to clarifying a child's knowledge. These ``developmental'' categories of negation do not emerge simultaneously; children tend to start using certain kinds before others \cite{nordmeyer2018individual}. 

Given that these categories represent some of the earliest uses of negation among humans, understanding how well TLMs can master them is important for building more human-like models of language processing. Understanding how well models perform on different categories will indicate whether they have mastery of some forms of negation, while also helping to identify failure points. Another interesting question is whether the proficiency of TLMs on these categories is at all related to competencies in human children (e.g., is the category which models consistently perform the best on the same that children most frequently employ?). However, to our knowledge, no prior work in NLP has focused on how well models perform on the forms of negation of interest to developmental psychology.

In this short paper, we investigate how well a suite of TLMs can process developmental negation,\footnote{By which we mean the forms of negation studied in development psychology.} by framing the problem as a natural language inference (NLI) task. We develop a rule-based parser to extract problems from existing NLI datasets, and evaluate our models on each category, in order to determine \textit{(i)} whether certain categories are more solvable by our models than others, and \textit{(ii)} what relationships exist among the categories. We find that models can consistently achieve stronger performance only on certain categories, and that training on combinations or sequences of these categories does not substantially improve a model's downstream performance.

\section{Related Work}

Negation is known to be frequently used in everyday conversation. While this includes its logical form, we primarily focus on negation's psycholinguistic forms, especially those that have been studied in the context of developmental psychology. Negation emerges early in child development, with `no' sometimes being a child's first word \cite{schneider2015large}, and even infants appear to understand forms of negation \cite{Piaget1980,HOCHMANN2021104599}. Preschool children use at least three different kinds of negation \cite{bloom1970language}, but possibly as many as nine \cite{choi1988semantic}. As noted by \citet{nordmeyer2018individual}, one of the first categories children use is \textit{rejection}, where a child rejects an object or activity. This is later followed by \textit{existence}, where a child might express the lack of an object, and later still \textit{denial}, which a child uses to deny the truth of a claim. Larger scale studies of child-directed speech have found that truth-functional kinds of negation tend to emerge later \cite{liu2021english}, but individual children do vary in their specific order of acquisition \cite{nordmeyer2018individual}. 
It is unknown whether this ordering reflects any deeper dependencies among the different categories, or whether the ordering is reflected in how artificial language models (LMs) learn negation. 

In NLP, methods from psycholinguistics have been used to probe the reasoning capabilities of LMs. Results from some studies have indicated that TLMs are not human-like in their processing of negation \cite{ettinger-2020-bert,kassner-schutze-2020-negated}. A similar line of work has used the NLI task to probe a model's ability to process negation and found that TLMs will often alter their predictions when negation is inserted or removed, even when the negation does not alter the entailment relationship \cite{hossain-etal-2020-analysis,hartmann-etal-2021-multilingual}.
As argued by \citet{kruszewski-etal-2016-logical}, part of the challenge of modeling purely logical negation is that a predicate often occurs in very similar contexts regardless of whether it is being negated. They argue that we should view negation as being a ``graded similarity function'', and show that distributional models can predict human plausibility judgments quite well, even in the presence of negation. These works show that it is unclear how well distributional models, especially TLMs, are actually processing negation. We contribute to this literature from a new perspective, by studying how well models can reason over forms of negation common in developmental psychology.

\section{The Developmental Negation Corpus}
\label{sec:corpus}

We use the NLI task to study the negation reasoning capabilities of our models. NLI problems consist of two sentences: a premise ($p$) and hypothesis ($h$), and solving such a problem involves assessing whether $p$ textually entails $h$. The generic structure of the NLI task makes it suitable for studying a variety of underlying reasoning skills, including negation. We specifically use the SNLI \cite{bowman-etal-2015-large} and MNLI \cite{williams-etal-2018-broad} datasets.

\begin{table}[tb]
\centering
\footnotesize
\begin{tabular}{lcc}
\hline
Category    & \# Train & \# Test \\ \hline
Possession ($PO$)  & 1053     & 520     \\ 
Existence ($EX$)  & 5528     & 2723     \\ 
Labeling ($L$)    & 2241     & 1104    \\ 
Prohibition ($PR$) & 814      & 400     \\ 
Inability  ($I$) & 1384     & 682      \\ 
Epistemic ($EP$)   & 1903     & 936     \\ 
Rejection   ($R$) & 1737     & 856     \\ \hline
\end{tabular}
\caption{Summary statistics for the curated dataset.}
\label{tab:dataset}
\end{table}

\begin{table*}[h]
\centering
\footnotesize
\begin{tabular}{lll}
\hline
    Category & Premise & Hypothesis \\
\hline
    $PO$ & \footnotesize{yeah you probably don't have the right temperatures...} & \footnotesize{You probably have ideal temperatures...} \\
    $EX$ & \footnotesize{This analysis pooled estimates...} & \footnotesize{The analysis proves that there is no link...} \\
    $L$ & \footnotesize{Not orders, no.} & \footnotesize{It is not orders.} \\
    $PR$ & \footnotesize{Two people are sitting against a building near shopping carts.} & \footnotesize{Run that way but don't run into the...} \\
    $I$ & \footnotesize{His manner was unfortunate, I observed thoughtfully.} & \footnotesize{I could not pick out what kind of manner he...} \\
    $EP$ & \footnotesize{yeah i don't know why} & \footnotesize{I know why} \\
    $R$ & \footnotesize{I lowered my voice...} & \footnotesize{I didn't want to be overheard.} \\
\hline
\end{tabular}
\caption{NLI examples extracted from each category, long examples have been trimmed to fit on one line.}
\label{tab:examples}
\end{table*}

To automatically identify questions that contain a specific kind of negation, we rely on the work by \citet{liu2021english} which studied how frequently different kinds of developmental negation occur in child-directed speech, using the data from the CHILDES corpus \cite{macwhinney2014childes}. To do this, they created a simple rule-based parser to automatically tag each sentence in CHILDES with the type of negation it contained (if any). We re-implement their parser, in some cases tweaking the rules slightly to better suit the structure of the NLI task. For each example across all the splits of both datasets, we first obtain a dependency parse of both $p$ and $h$ using the diaparser package \cite{wang-etal-2019-second}, and check if either contains an explicit negation marker (``no'', ``not'', or ``n't''). If one span contains negation, we check if the syntactic structure obeys the rules of any of our categories. If the span falls into a category, we mark it as belonging to that category. We use these questions as the diagnostic set for our experiments, splitting out 1/3 of the questions in each category as a \textit{diagnostic test} set, and leaving the remainder as a \textit{diagnostic train} set (and we will refer to them as such). We place the remaining NLI questions containing no negation in a separate $NLI_{train}$ set, giving us about 730,000 examples we use to finetune our models on the NLI task. We split out 9,000 questions from this train set at random to use as a $NLI_{dev}$ set, balanced for each label. In the following, we describe the precise rules used to determine which category a negated example should be assigned to:

\paragraph{Possession ($PO$)} We require that the lemma of the root be \textit{have}, \textit{has}, or \textit{had}, and that the root is directly modified by both the negation and the verb \textit{do}.

\paragraph{Existence ($EX$)} We require that \textit{there} occur in the text and precede the negative marker and that the negative marker directly modifies a noun phrase, determiner, or an adverb.

\paragraph{Labeling ($L$)} We require that the sentence begin with either \textit{That} or \textit{It}, and that the root of the sentence is a noun which is modified by \textit{is} or \textit{'s}.

\paragraph{Prohibition ($PR$)} We require that the sentence not contain a subject and that the negation is immediately preceded by \textit{do}. To not conflate this category with others, we filter out cases where the root contains one of the explicit markers of another category (e.g., \textit{like} or \textit{want} in the case of rejection).

\paragraph{Inability ($I$)} We require that the negation directly modify the root of the sentence, and that the word immediately before the negation is either \textit{can} or \textit{could} (e.g., \textit{can not do}). Prior literature has typically viewed inability from an egocentric perspective. However, we found that allowing only the first person severely restricted the number of examples extracted, and therefore chose to also allow the second and third person.

\paragraph{Epistemic ($EP$)} We require that the root be \textit{remember}, \textit{know}, or \textit{think}, and that the root be directly modified by the verb \textit{do}.

\paragraph{Rejection ($R$)} We require that the lemma of the root word be either \textit{like} or \textit{want}, and that the root is modified by the negative marker.

After performing extraction, categories $L$ and $PR$ contained fewer than 1000 examples, which we deemed was insufficient to split into separate train and test sets. To address this, we developed a simple data augmentation approach that utilized the Wordnet database \cite{miller1998wordnet}. From the dependency parse of both $p$ and $h$, we check if the root of either parse occurs in both spans. If it does, we obtain all synonyms of the word in Wordnet and replace the root in both spans with the synonym (doing this for every synonym). We found this simple approach increased the number of examples for both $L$ and $PR$ to at least 1500. Note that we performed no augmentation for the other categories, as our parser extracted at least 1500 examples for all other cases. Table \ref{tab:dataset} shows statistics for the dataset after augmentation.

Table \ref{tab:examples} shows extracted examples, along with their category assignment. We generally found that the extracted examples matched up with the prototypical category quite well, although in some cases their semantics differed slightly. For instance, consider a $PR$ example with $p$ = \textit{don't miss having a flick through the albums} and $h$ = \textit{The pictures of old Madeira show a more interesting city than now}, which is an MNLI example originally extracted from a travel guide. Although this technically counts as $PR$, it does not have quite the same semantics as an actual command. Unfortunately, these ambiguities are not easily resolved, given that negation takes on many forms and may occur at any location within a sentence. We, therefore, opted to focus on forms of negation that can be easily extracted, and leave improvements to our dataset creation protocol for future work.

\section{Experiments}

Using the curated dataset, we performed a series of exploratory experiments to help us understand how well TLMs process each of the negation categories. We use BERT \cite{devlin-etal-2019-bert}, and RoBERTa \cite{liu2019roberta}, two popular transformer LMs that have demonstrated impressive results on a variety of language understanding tasks. We also examine MiniBERTa \cite{warstadt-etal-2020-learning} and BabyBERTa \cite{huebner-etal-2021-babyberta}, which are both based on the RoBERTa architecture but were pre-trained on a much smaller number of tokens (10 million and 5 million respectively), which is more realistic to the amount of language a child is exposed to in the first few years of life. We use the Huggingface implementation of all models \cite{wolf-etal-2020-transformers}, and use both the \textit{base} and \textit{large} version of BERT and RoBERTa, which differ only in the number of trainable parameters.

\begin{figure}[h!]
    \centering
    \includegraphics[width=1\linewidth]{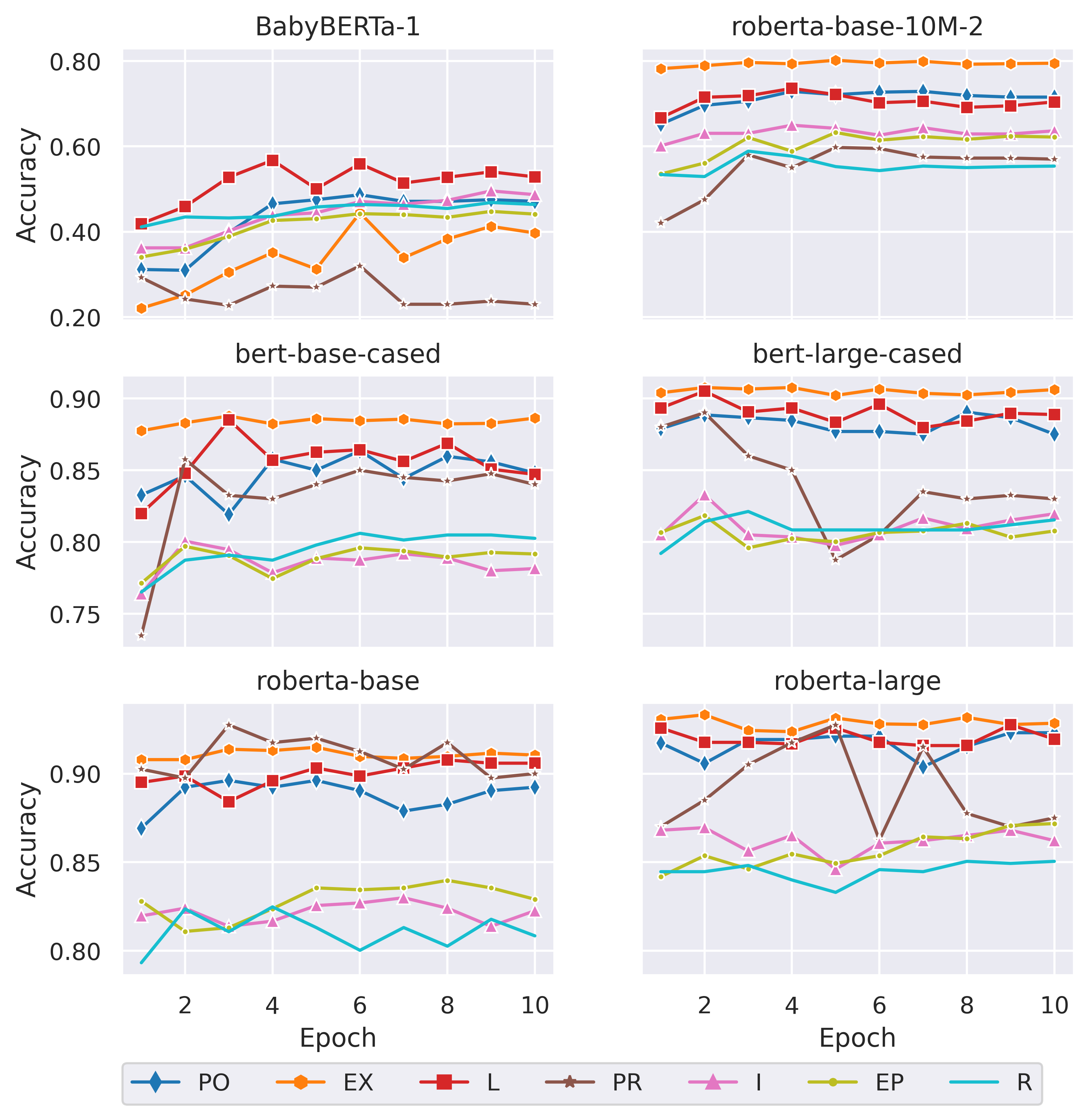}
    \caption{Performance of models finetuned on $NLI_{train}$ for each diagnostic test set. We refer to MiniBERTa using its Huggingface model ID (\textit{roberta-base-10M-2}).}
    \label{fig:experiment_1}
\end{figure}

\paragraph{Experiment 1:} We began by investigating whether TLMs would master certain negation categories sooner than others over the course of training. We train our models on $NLI_{train}$ for 10 epochs, using a learning rate of $1e-5$, a weight decay of $0.01$, a batch size of $16$, and a maximum sequence length $175$.\footnote{We set the maximum sequence length for BabyBERTa to 128, which is the longest that the model supports.} We selected these hyperparameters to be similar to those which were previously reported to yield strong results when training on NLI datasets \cite{laverghetta-jr-etal-2021-transformer}. We additionally evaluated the models on $NLI_{dev}$, and found that they all achieved a Matthews Correlation of at least 0.6 \cite{matthews1975comparison}, and thus concluded that these hyperparameters were suitable. For every end of epoch checkpoint across all models, we obtained evaluation results on each diagnostic test set. Importantly, the models are not finetuned on any negated NLI questions for this experiment, meaning that all knowledge of negation comes from pre-training. Results are shown in Figure \ref{fig:experiment_1}. We see that the categories have similar rankings in terms of accuracy. For example, $L$ and $PO$ are among the top two best-performing categories, while $R$ is generally one of the worst-performing ones, indicating clear distinctions in how LMs process the categories. BabyBERTa, unlike other models, also shows stronger similarities to how children acquire negation. For instance, while $R$ is thought to be one of the first categories children acquire, BabyBERTa is the only model where $R$ is one of the highest-ranking categories in terms of accuracy.

\begin{figure}[h!]
    \centering
    \includegraphics[width=1\linewidth]{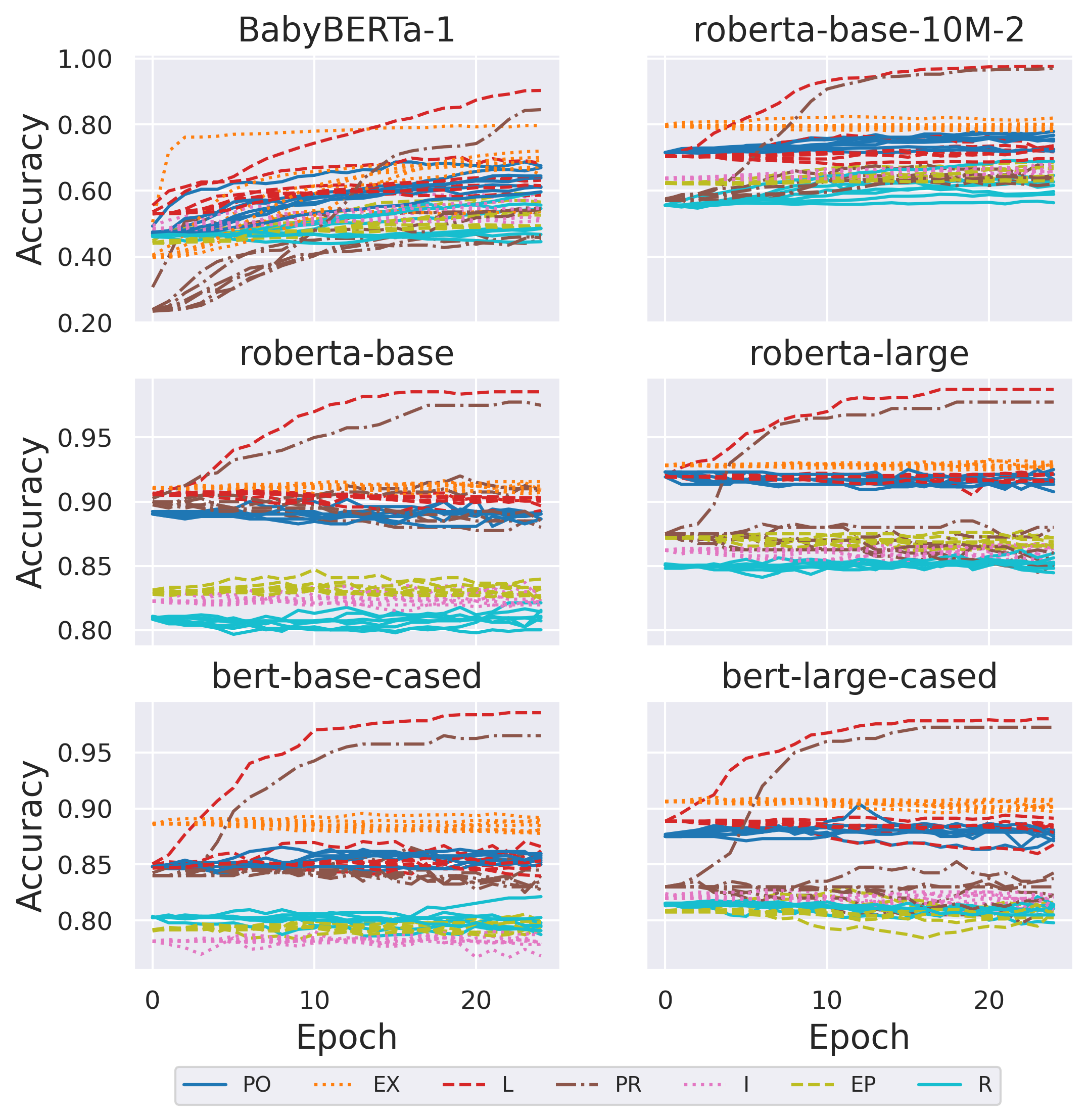}
    \caption{Accuracy of each model on every diagnostic test set, after being finetuned on every diagnostic train set. Plots are color-coded based on the target category.}
    \label{fig:experiment_2}
\end{figure}

\paragraph{Experiment 2:} One might expect that children develop a more abstract understanding of negation as they are exposed to different categories. This was suggested by \citet{pea1978development} who argued that more abstract forms of negation develop from less abstract ones, suggesting that mastering one form of negation can lead to positive transfer on others. In Experiment 2, we examined how much positive transfer could be obtained from training on one of the negation categories, and then testing on the others. We adopt a similar methodology to \citet{pruksachatkun-etal-2020-intermediate}, who explored the conditions that affect intermediate task transfer learning. Using the models trained in Experiment 1, we further finetune these models for 25 epochs on each diagnostic train set separately. We then evaluate the finetuned models on each diagnostic test set, which allows us to examine all possible pairwise interactions among categories.  Figure \ref{fig:experiment_2} shows the results for all combinations of diagnostic categories for training and testing. Surprisingly, we find that positive transfer generally only occurs when a model is trained on the same category it is being tested on. Training on a different category has little to no effect on the target category. BabyBERTa is again an exception, as we do see positive transfer for most pairs, suggesting the model is generalizing across categories

\begin{figure}[h!]
    \centering
    \includegraphics[width=1\linewidth]{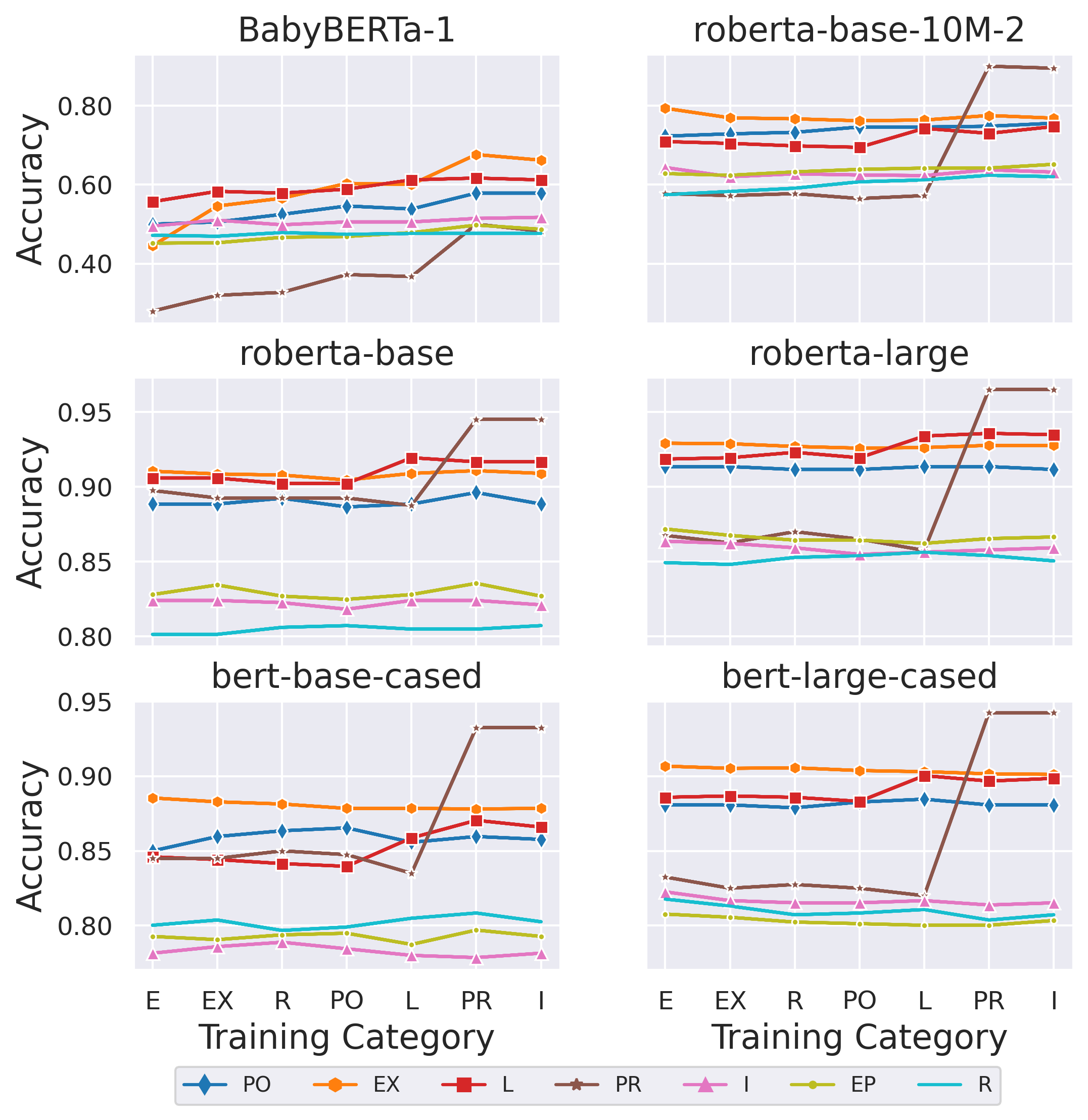}
    \caption{Results from Experiment 3. The x-axis shows the sequence of categories on which all models were trained, while the y-axis shows the accuracy obtained after being trained on a category.}
    \label{fig:experiment_3}
\end{figure}

\paragraph{Experiment 3:} Building on Experiment 2, we examined how the performance of our models is affected when trained on all diagnostic categories in sequence. Assuming that no positive transfer exists among the categories, we would expect to see a model's performance on a particular category improve only after it has been trained on that same category, and even training on multiple other categories should not substantially improve performance on the target. Using the models from Experiment 1, we finetune each model for 10 epochs on every diagnostic train set, using the sequence of categories shown in the x-axis of Figure \ref{fig:experiment_3}. Additionally, we under-sample all diagnostic train sets to have the same number of questions as $PR$, so that all categories contribute the same amount of data. Figure \ref{fig:experiment_3} shows the results. For some categories, such as $L$ and $PR$, we see the expected trend. The largest accuracy gain for these categories occurs whenever the model is trained on the same category it is being tested on, and performance drops slightly after being trained on others. However, for categories such as $R$, the best performance gain is not always after being trained on the same category. We sometimes see the model continue to improve on $R$ after being trained on $R$, and in some cases, training on $R$ causes performance on $R$ to \textit{decrease}.

\section{Discussion and Conclusion}

In this paper, we have explored how well transformers process categories of developmental negation. We find that performance rankings across categories are generally consistent, but that the categories seem to test for orthogonal skills in the majority of LMs. In BabyBERTa, we see significant similarities with the order of negation acquisition in children. Two of the best performing categories are $R$ and $L$, while two of the worst are $EX$ and $PR$, which aligns quite well to the order observed by \citet{liu2021english}. It thus seems that TLMs do at least partially reflect the order of negation acquisition observed in children, although more experiments would be needed to understand the extent of this correlation. That we found category rankings to generally be consistent across LMs may have interesting implications, and understanding why LMs struggle with certain categories may help to improve the ability of LMs to process negation.

Future work can build on these experiments in several ways. In Experiments 2 and 3, we modeled interactions among the negation categories in either a pairwise or sequential fashion, which is unlikely to reflect how children are exposed to negation. More experiments, mixing all of the categories at once in various proportions, might yield a more realistic model of cognitive development. Our approach also requires that each category fits into a specific structure, which limits the amount of examples that can be extracted. Future work will need to expand our ruleset to include more variations in the negated utterances covered. Finally, while we primarily focus on finetuning, pre-training is likely to impact the proficiency of our models on the categories as well. Future work should precisely control the prevalence of each category in the pre-training corpus, to observe what effect this has on downstream performance.

\bibliography{Antonio}
\bibliographystyle{acl_natbib}

\end{document}